# BD-SHS: A Benchmark Dataset for Learning to Detect Online Bangla Hate Speech in Different Social Contexts

**Nauros Romim[1], Mosahed Ahmed[2], Md. Saiful Islam[3]**
**Arnab Sen Sharma[4], Hriteshwar Talukder[5], Mohammad Ruhul Amin[6]**
Shahjalal University of Science and Technology[1, 2, 3, 4]
University of Alberta, Canada[3]
University of Boston, USA[5]
Fordham University, USA [6]
naurosromim@gmail.com[1], mosahed32@student.sust.edu[2], saiful-cse@sust.edu[3]
arnab-cse@sust.edu[4] hriteshwar-eee@sust.edu[5] mamin17@fordham.edu[6]

**Abstract**

Social media platforms and online streaming services have spawned a new breed of Hate Speech (**HS**). Due to the massive amount of user-generated content on these sites, modern machine learning techniques are found to be feasible and cost-effective to tackle this problem. However, linguistically diverse datasets covering different social contexts in which offensive language is typically used are required to train generalizable models. In this paper, we identify the shortcomings of existing Bangla HS datasets and introduce a large manually labeled dataset **BD-SHS** that includes HS in different social contexts. The labeling criteria were prepared following a hierarchical annotation process, which is the first of its kind in Bangla HS to the best of our knowledge. The dataset includes more than 50,200 offensive comments crawled from online social networking sites and is at least 60% larger than any existing Bangla HS datasets. We present the benchmark result of our dataset by training different NLP models resulting in the best one achieving an F1-score of 91.0%. In our experiments, we found that a word embedding trained exclusively using 1.47 million comments from social media and streaming sites consistently resulted in better modeling of HS detection in comparison to other pre-trained embeddings. Our dataset and all accompanying codes is publicly available at `github.com/naurosromim/hate-speech-dataset-for-Bengali-social-media`

## 1. Introduction

Social media and online streaming platforms are gradually becoming an integral part of peoples' daily lives. These platforms have democratized information and allowed people a greater freedom of speech, sometimes even offering anonymity. However, as a result, these sites have become a new frontier for spreading misinformation and HS. But the anonymous and global nature of these sites makes it difficult for law enforcement agencies to regulate them (Gagliardone et al., 2015). Identifying HS and curtailing its spread is crucial to safeguard human rights and prevent it from marginalizing individuals and groups of people based on their race, gender, ethnicity, or other affiliations. And, due to the massive amount of user-generated data on these kinds of sites, the only feasible solution to effectively tackle this challenge is by utilizing modern natural language processing and machine learning models. However, the lack of formal language syntax, spelling mistakes, and use of various swear and non-standard acronyms in the comment sections of social media and online streaming sites makes this task harder (Nobata et al., 2016). Researchers have started working in this very challenging domain by preparing large, linguistically diverse gold standard datasets in order to train machine learning classifiers for the task. However, the majority of these efforts have been focused on English language (Schmidt and Wiegand, 2017) and low resource languages like Bangla are lagging behind. *Hatespeechdata*[1] is a website that tracks the development HS datasets across different languages. At the time of this writing, HS datasets in 13 different non-English languages are listed on this site.

There are close to 46 million *Facebook*[2] users and 29 million *Youtube*[3] users in Bangladesh. Despite this large userbase, there has been a severe lack of large, linguistically diverse Bangla HS datasets. We compared existing English and other language HS dataset with existing Bangla HS datasets and identified some critical limitations.

### 1.1. Limitations of Existing Bangla HS Datasets

**Dataset size and linguistic diversity: Table 1** depicts an overview of currently available Bangla HS datasets. We can see that the majority of these datasets contain less than 10k HS entries. In order to increase the ratio of HS, (Waseem and Hovy, 2016) suggested extracting comments from contentious topics which are more likely to contain high concentrations of HS, such as *Islam terror*. But (Schmidt and Wiegand, 2017) cautioned that such strategy would most likely make the dataset oriented towards specific topics and hence reduce the linguistic diversity. As a consequence, machine learning classifiers trained on such datasets would fail to generalize properly.

[1] https://hatespeechdata.com/

[2] https://www.statista.com/statistics/268136/top-15-countries-based-on-number-of-facebook-users/

[3] https://www.statista.com/forecasts/1146236/youtube-users-in-bangladesh

| Dataset | Dataset Size | No of HS | Agreement Score | HS Target |
|---|---|---|---|---|
| (Chakraborty and Seddiqui, 2019) | 5,644 | 2,500 | No | No |
| (Emon et al., 2019) | 4,700 | 3,137 | No | No |
| (Awal et al., 2018) | 2,665 | 1,214 | No | No |
| (Ishmam and Sharmin, 2019) | 5,126 | 3,178 | No | No |
| (Banik, 2019) | 10,219 | 4,255 | No | No |
| (Karim et al., 2020a) | 6,115 | 6,115 | Yes | No |
| (Sharif and Hoque, 2021) | 14,158 | 6,807 | Yes | No |
| (Romim et al., 2021) | 30,000 | 10,000 | No | No |
| **BD-SHS** | **50,281** | **24,156** | **Yes** | **Yes** |

Table 1: Overview of Bangla HS datasets

**Lack of detailed annotation guideline:** Annotating HS is inherently a complex and challenging task(Nobata et al., 2016). Waseem (2016) showed how an annotator's bias can influence the annotation process and, subsequently, affect the classification task. To combat this, de Gibert et al. (2018) prepared a stringent and detailed annotation guideline, providing specific points on what constitutes HS and what does not. They also claim that such guidelines improve the consistency and reliability of annotation by reducing subtle biases among the annotators. However, to the best of our knowledge, amongst Bangla HS datasets, only (Romim et al., 2021) and (Sharif and Hoque, 2021) followed such detailed HS guidelines.

**Lack of annotator's agreement score:** Annotator agreement score is an important metric for understanding the dataset's annotation quality. But so far, only (Karim et al., 2020a) and (Sharif and Hoque, 2021) had any kind of annotator agreement score. Which makes the reliability of most of the existing Bangla HS datasets questionable.

**Lack of hierarchical annotation scheme:** (Zampieri et al., 2019a) suggested the need for an annotation scheme distinguishes among the *target*s of HS in order to get better insights into the nature of HS. (Sigurbergsson and Derczynski, 2020) (Danish) (Mubarak et al., 2020) (Arabic) (Pitenis et al., 2020) (Greek) (Çöltekin, 2020) (Turkish) have annotated their datasets to classify the target of offensive comment into individual and group. But to the best of our knowledge, similar annotation schemes remain unexplored in currently available Bangla HS dataset.

There are some existing Bangla HS dataset that have categorized HS into different types. (Karim et al., 2020a) and (Sharif and Hoque, 2021) categorized HS into four types. However, these studies did not go any further to identify the target victims of certain HS.

### 1.2. Contribution of This Paper

To overcome these shortcomings of Bangla HS datasets discussed in the previous section, we present the **BD-SHS**, a large, manually annotated dataset of Bangla HS in a variety of social contexts. We have collected 50,281 comments from the comment sections of different social media sites and carefully annotated them following a three-level hierarchical annotation scheme in order to analyze and understand HS in different social contexts.

- **Level 1:** HS Identification.
- **Level 2:** Identification of the Target of HS.
- **Level 3:** Categorization of HS Types.

Among the 50,281 comments, 24,156 are tagged as HS. To the best of our knowledge, this is the largest Bangla HS dataset currently available, at least 50% larger than the previous largest, publicly available Bangla HS dataset.

We prepared and followed a strict annotation guideline to reduce human annotator bias. We present the annotator agreement score of our dataset. We benchmark our dataset by experimenting with different machine learning models and linguistic features.

## 2. Dataset

### 2.1. Data Collection

Our dataset is comprised of comments collected from online social media and online streaming sites. Due to the fact that Bangladeshi internet users are still relatively inactive in the comment sections of blogs and online news portals, these sites were excluded from this study. To increase our chances of obtaining HS comments, we began by curating a list of controversial events that occurred in Bangladesh following 2017. We do not get back further than 2017 because we noticed controversial posts or videos prior to that year received little engagement in their comment sections. To curate this list of controversial topics, we conducted a short survey of 20 undergraduate students (12 males and 8 females) from 17 different majors of Shahjalal University of Science and Technology. All of participants were between the ages of 20 and 24, and frequent users of online social media and video streaming platforms. We briefed them on our objective and asked them to submit individually a list of contentious issues and keywords for us to consider as we prepared our HS dataset. To ensure linguistic diversity, our participants were asked to suggest topics or keywords from a range of domains. To determine whether the topics are indeed from diverse domains, we analyzed the topics and were able to cluster them into broadly six categories: *sports*, *entertainment*, *crime*, *politics*, *religion*, and *controversial comments by scholars' (CCS)*.

| # | Comment | HS | Target | Type |
|---|---|---|---|---|
| 1 | কি বালের মুভি<br>What an ass movie. | NH | - | - |
| 2 | বোরকা পরে না, ধর্ষিত তো হবেই।<br>It is obvious that girls who do not wear burqas would be raped. | HS | female | gender |
| 3 | তর কাজ দেখে বমি করে দিতে ইচ্ছে করছে।<br>Seeing your work makes me want to puke.” | HS | IND | slander |
| 4 | দাড়ি ওয়ালা তো এক নাম্বার বাইনচোত আর বাকি গুলা রে জুতা মারা দরকার<br>That bearded guy is number one asshole and the rest should be beaten with shoes. | HS | male, group | slander, CV |
| 5 | তুই কুত্তার বচ্চা। তোর মত মালাউনের দেশে থাকার কোন অধিকার নাই। তোরে আমি পিটায়ে মেরে ফালাম।<br>You son of a bitch. A *malaun* (*swear for Hindu*) like you has no right to live in our country. I will beat you to death. | HS | IND, group | slander, religion, CV |
| 6 | অডমিন সালা তুমি কই সালা পাগল কই থেকে আসে<br>Admin, where are you bastard? Where does these kinds of lunatics come from? | HS | IND | slander |

Table 2: Representative snapshot of our dataset

Then, we searched based on these keywords for publicly available contents on social media sites (such as posts and pages on *Facebook*) as well as online streaming platforms (such as *Youtube* videos) and scraped the entire comment sections using an open source tool called *Facepager* [4]. Additionally, we searched for Bangla *TikTok*, roasting, and similar videos, as we observed that the comment section in these videos tend to be very toxic. We categorized comments scraped from such videos as *miscellaneous*. All of the commentators' names, affiliations, and other information were excluded from our dataset to ensure anonymity.

We collected more than 100k comments in total. Then, using a `Jaccard Index` cutoff of **0.8**, we removed duplicate and highly similar comments to ensure diverse vocabulary. Finally, we had **50,281** comments remaining for our annotation process.

[4]https://github.com/strohne/Facepager

## 2.2. Overview of Annotation Scheme

The following subsections discuss our level-wise annotation process in detail.

### 2.2.1. Level 1: HS Identification

Inspired by (Nobata et al., 2016), this level focuses on binary classification between **HS** and Not Hate (**NH**). The criteria for both classes are discussed below.

A comment is annotated as ***HS***, if:

- It attacks a group or individual based on ethnicity, nationality, religion, sexual orientation, age, gender, physical and mental disability, and disease.
- It does not directly dehumanize a person but supports the act of HS. For example, in the second comment in **Table 2**, the commentator did not intend to rape a person but supports the act of rape.
- It expresses the intention of inflicting violence or supporting the act of violence against a person or community.
- It dehumanizes a group and/or an individual by comparing them with insects, animals, known criminals, or historically villainous person.
- It expresses disgust for another person. The third comment in **Table 2** fits this description.

A comment is annotated as ***NH***, if:

- It does not dehumanize or attack an individual or group through the above discussed means.
- If a sentence contains a swear word but is not attacked towards a human being, it is not HS. For first comment in **Table 2**, the swear word was attacked towards the movie, not the director or the actors. So it was not considered as HS.

#### 2.2.2. Level 2: Identification of the Target of HS

Level 2 focuses on annotating the targets of the hate speach comments. Only the comments annotated as **HS** in Level 1 of annotation are considered in Level 2. The classes of Level 2 annotation are below.

- **Individual (IND):** HS comments target an individual person, but it is not clear from the language if the targeted person is male or female.
- **Male:** It is clear that the target of HS comment is a male.
- **Female:** It is the clear that the target of HS comment is a female.
- **Group:** HS comments target a group of people considered as unity due to their same ethnicity, religious belief, gender, political affiliation, an institution, or other common characteristics.

We treated this as a multilabel task as it is possible for a hate speech comment to target multiple individuals, or an individual and a group in the same comment. Example 4 of **Table 2** threatens one individual while insulting followers of a religion at the same time. So, the target of the comment is both an *individual* and a *group*.

#### 2.2.3. Level 3: Categorization of HS Types

In Level 3, we focused on determining the types of offense in HS comments. Our selected labels are below.

A comment is annotated as:

- **Slander:** If the comment targets an individual or group using swear or cursed words, or dehumanises comparing them to animal or insect, etc.
- **Call to violence (CV):** If the comment expresses the intention of inflicting violence on a group or individual, or supports such action.
- **Gender:** If the comment targets a person or group based on cis or preferred gender identity.
- **Religion:** If the comment targets an individual or group based on their religious beliefs.

Same as Level 2, Level 3 is also treated as a *multilabel* task. In example 5 of **Table 2**, two distinct types of curse words were used. One is a common profane expression, while the other is directed specifically at followers of *Hinduism*, and then it expresses the intention of inflicting violence on the targeted individual. As a result, this comment has been annotated as *slander*, *religious*, and *call to violence* (**CV**).

### 2.3. Annotation

We prepared the annotation guideline gradually in multiple steps. First, we randomly sampled 1000 comments. Then, for Level 1, we prepared a draft of the annotation guideline based on community standard of *Facebook* and *YouTube*. Following that, we conducted a trial annotation by annotating the selected 1000 comments. There were numerous comments that caused confusion among our annotators regarding their proper label. We discussed over those comments, examined the community guidelines more closely, and finally crafted the final annotation guideline for this level.

Level 2 was inspired by *OffensEval*'s Sub-task 3 by (Zampieri et al., 2019b), which categorized HS targets into 2 classes: *ind, group*. However, during annotation, we discovered that for some HS comments it was not difficult to identify the targets as *male* or *female*. So, we decided to categorize the targets of HS comments into 4 classes: *ind*, *male*, *female*, and *group*.

Level 3 was inspired by (Ibrohim and Budi, 2019) which categorized HS into *religion, racism, physical disability, slander, gender*. However, in our trial annotation, we found that physical disability and slander are often indistinguishable. For example, in example 6 of **Table 2**, the word *lunatics* is used as slander. Since the number of HS comments regarding race are rare and more often they are used to demean non-muslim ethnic groups, we consider the offense type of such comment types as *religious*. (Sharif and Hoque, 2021) also mentioned similar observation while preparing their *Bengali aggressive text* dataset. We also found that numerous comments intending to inflict violence or supports these acts of violence and tag such comments as *call to violence* or **CV**. So, we finally settled on 4 HS types: *religion*, *slander*, *gender*, and *CV*. We finalized a detailed annotation guideline with examples for all 3 levels for final annotation phase, which was explained in section 2.2.

For the annotation process, 50 undergraduate students (32 males and 18 females) were recruited from 20 different majors of Shahjalal University of Science and Technology. They are frequent users of social media platforms and online streaming sites.

We briefs them of the sensitive nature of the task and they willingly volunteered to participate in this study. We conducted a 3-hour long training session with our annotators, briefing them on our annotation guidelines prior to starting the annotation process.

For Level 1, each of the comments was annotated by three annotators, and the majority decision was taken as the final decision. The inter-annotator agreement, `Fleiss Kappa` score of `0.658`, indicates moderate agreement among our annotators.

### 2.4. Dataset Statistics

A statistical breakdown of our dataset is shown in **Table 3**. There is a total of 24,156 HS comments in our dataset, which constitutes roughly 48% of the total. So, Level 1 of our dataset (*Hate Speech Identification*) is almost balanced.

A noteworthy observation of Level 3 is that, the number of *slander* comments is disproportionately larger than other labels. Consider comments 4 and 5 of **Table 2**. Each of these comments express *slander* along with other types of offense such as *call to violence* or *religion*. We have also observed that a large number of comments simply express hatred towards specific individuals or groups, but do not elaborate on the cause of this hatred.

**Table 4** shows the distribution of the association between the *target*s of HS with the *type*s of HS. From **Table 4** it is apparent that *female*s are disproportionately targeted on the basis of gender. On the other hand, *female*s are less likely to be targeted on the basis of just *slander*. It implies that when a female individual is the target of HS, it is more likely that the reason is simply her being a female. It can also be observed that, *group* is less subjected to *call to violence* type HS compared to other targets. It indicates that people who write HS comments are more likely to express the intention of inflicting violence upon individuals rather than on a group or community.

Our dataset is noisy; majority of the comments contain mixed dialacts as well as grammatical and spelling errors. A curse word can have multiple variations due to spelling mistakes. Additionally, we observed several comments with unfinished sentences, multiple words forming a single word due to lack of white space and other noises during the annotation process.

### 2.5. Multi-purposeness of our dataset

Our dataest can be used to tackle multiple classification tasks. Those are discussed below.

- **Cyberbullying**: Cyberbullying are insults or derogatory comments targeted towards individuals which propagate through digital media (often social media platforms) (Zampieri et al., 2019b). Level 2 of our annotation scheme intends to compile a dataset capable of identifying the *target*s of HS, thereby, our dataset has potential application in training cyberbullying detection models and analyzing cyberbullying languages in social media platforms.
- **Aggression identification**: In Level 3 of our annotation scheme we identify HS comments that expresses explicit intention of inflicting violence upon individuals or groups and annotate them as *call to violence*. So, our data has potential applications in filtering such comments and ensuring a safe, inclusive space in social media platforms.

| Level | Annotation labels | # comments |
|---|---|---|
| Level 1 | HS | 24,156 |
| | NH | 26,125 |
| Level 2 | Ind | 8,815 |
| | Male | 7,128 |
| | Female | 5,443 |
| | Group | 3,119 |
| Level 3 | Slander | 16,992 |
| | Religion | 1,562 |
| | Call to violence | 7,232 |
| | Gender | 4,244 |

Table 3: Distribution of annotated comments across three levels.

| HS Targets | Slander | Religion | Gender | CV |
|---|---|---|---|---|
| Ind | 80% | 7% | 4% | 33% |
| Male | 88% | 4% | 3.5% | 32% |
| Female | 30% | 3% | 65% | 38% |
| Group | 73% | 18% | 7% | 25% |

Table 4: Distribution of HS types among HS targets

## 3. Experiment

After performing a random shuffle, the dataset was split into **train (`70%`)**, **validation (`15%`)**, and **test (`15%`)** sets. We experimented with various models and linguistic features to develop benchmark results for three classification tasks permitted in our dataset:

1. **Task A:** HS Identification (*Binary classification*)
2. **Task B:** Identify the Target of HS (*Multi-label classification*)
3. **Task C:** Categorization of HS Types (*Multi-label classification*)

### 3.1. Linguistic Features

**Lexical:** We experimented with word unigrams and character *ngrams* of length 2 to 5. We calcu-lated the *Term Frequency Inverse Document Frequency* (**TF-IDF**) weighted scores for each of the word and char *ngrams*, and then fed these TF-IDF scores as features to classifiers. In case of words, we show results of only *unigrams* because combining *unigram* with *bigram* and *trigram*, or using only *bigram* or *trigram*, resulted in poor F1 scores during the validation phase.

**Pre-trained word embedding:** We experimented with *BengFastText* (**BFT**) (Karim et al., 2020b), which is pretrained on 250 million Bangla articles, and multilingual fasttext (**MFT**)(Grave et al., 2018), pretrained on articles in 157 languages. It is worth noting that both of these models are pretrained on *Wikipedia*[5] articles.

Khondoker Ittehadul Islam (2021) explained that the Bangla text available in Wikipedia is *formal* i.e contains few spelling and grammatical mistakes and mixed dialects. On the other hand, social media comments tend to be *informal* or more noisy in terms of more spelling and grammatical mistakes and contains mixed dialects. They showed in their *noisy Bangla sentiment dataset* that word embeddings trained on formal text performed poorly on informal social media data.

Our dataset is also comprised of noisy and *informal* social media comments. So, we wanted to check if this observation is consistant is our dataset as well. We further wanted to examine if a word embedding trained with *noisy*, *informal* texts can perform relatively better is modeling HS in our dataset. So, we collected 1.47 million Bangla comments from the comment sections of *Facebook* and *Youtube*. While preparing the training corpus for this word embedding model, we ensured language diversity by collecting comments from posts or videos from eight different categories: *education, entertainment, health, influencer, religion, politics, sports,* and *technology*. We introduce an acronym **IFT** i.e. **I**n**F**ormal **T**ext, to indicate embeddings extracted using this *FastText* language model trained from scratch with informal texts.

### 3.2. Models

**SVM:** We trained linear Support Vector Machine (SVM) (Cortes and Vapnik, 1995) with TF-IDF weighted scores of different combinations of word and character *ngrams*. Hyperparameters such as *regularizer C*, *penalty*, and *loss* were finetuned to find the best performing combinations on the validation set. For multilabel classification task, we used the *one-vs-rest* strategy.

**Bi-LSTM:** We experimented with Bidirectional Long Short Term Memory (Bi-LSTM) architectures (Hochreiter and Schmidhuber, 1997). Since Bi-LSTM architectures are capable of taking sequences of arbitrary lengths as input, we fed into this architecture word *embedding vectors* of each of the words in a comment one by one as a *sequence*. We obtained the *embedding vector* of an word from **BFT** and **MFT**, trained with formal texts and **IFT**, the language model we trained with informal texts. We also used an updatable embedding layer learned by the model during training. From now on, it is denoted as *random embedding* (**RE**). The Bi-LSTM architecture we experimented with consists of i) a bidirectional LSTM layer of 100 units, ii) an average pooling layer, iii) a fully connected hidden layer of 16 units, and iv) 2 output nodes with *softmax* activation function for taskA and 4 output nodes with *sigmoid* activation function for taskB and taskC.

[5]https://www.wikipedia.org/

## 4. Result

In this section, we present the results we obtained for each of the three sub-tasks by training our classifier models with different combinations of features.

### 4.1. Task A: HS Identification

The result is given in **Table 5**. BiLSTM trained with informal embeddings from **IFT** achieved the highest `F1 score` of **91.0%**. Note that, BiLSTM with informal embedding (**IFT**) outperformed both formal embeddings (**BFT** and **MFT**). SVM trained with char *ngrams* is comparable, with an `F1 score` of **90.9%**. (Schmidt and Wiegand, 2017) explained that char ngram is very useful to attenuate the spelling variation problem.

### 4.2. Task B: Identify the Target of HS

The results for this sub-task are presented in **Table 6**. Since we did not have similar number of HS comments in each of the targets, we present the *Weighted Average* (**WA**) of each types of scores on the last column of **Table 6** in order to give an idea of the overall performance of the models. Similarly sub-task A, BiLSTM trained with **IFT** outshines both of the formal embeddings (**BFT** and **MFT**). BiLSTM trained with **IFT** achieved the best weighted average `F1 score` of **77.5%**. However, in sub-task B, SVM trained with char *ngrams* (**SVM + C**) failed to follow up on its impressive performance in sub-task A.

From average recall, we can deduce that models had the most difficulty correctly predicting *group* and *individual* as the targets of HS.

### 4.3. Task C: Categorization of HS Types

**Table 7** depicts the results for this sub-task. SVM trained with char *ngrams* (**SVM + C**) achieved

| Model + Feature | HS | | | NH | | | Weighted Average | | |
|---|---|---|---|---|---|---|---|---|---|
| | P | R | F1 | P | R | F1 | P | R | F1 |
| SVM + U | 91.4 | 84.4 | 87.7 | 86.5 | 92.7 | 89.5 | 88.9 | 88.7 | 88.7 |
| BiLSTM + MFT | 90.2 | 86.9 | 88.6 | 88.3 | 91.3 | 89.8 | 89.2 | 89.2 | 89.1 |
| BiLSTM + BFT | 89.6 | 89.3 | 89.4 | 90.1 | 90.4 | 90.3 | 89.9 | 89.9 | 89.9 |
| BiLSTM + RE | 90.6 | 88.5 | 89.5 | 89.6 | 91.5 | 90.5 | 90.1 | 90.0 | 90.0 |
| SVM + U + C | 90.1 | 92.4 | 91.3 | 91.5 | 89.1 | 90.3 | 90.8 | 90.7 | 90.8 |
| SVM + C | 92.1 | 88.8 | 90.4 | 89.9 | 92.9 | 91.5 | 91.1 | 90.9 | 90.9 |
| BiLSTM + IFT | 90.8 | 90.5 | 90.7 | 91.2 | 91.5 | 91.4 | 91.0 | 91.0 | 91.0 |
| Average | 90.7 | 88.7 | 89.7 | 89.6 | 91.4 | 90.5 | 90.1 | 90.1 | 90.1 |

Table 5: Result for Task A: HS identification. Results are sorted in ascending order based on weighted F1. **U**: word unigram, **C**: char *ngrams*

| Model + Feature | IND | | | Male | | | Female | | | Group | | | Weighted Average | | |
|---|---|---|---|---|---|---|---|---|---|---|---|---|---|---|---|
| | P | R | F1 | P | R | F1 | P | R | F1 | P | R | F1 | P | R | F1 |
| BiLSTM + BFT | 70.2 | 54.9 | 61.6 | 71.6 | 58.4 | 58.4 | 71.5 | 68.6 | 70.1 | 71.1 | 46.8 | 56.4 | 71.0 | 58.1 | 63.7 |
| SVM + U | 71.7 | 63.8 | 67.6 | 78.8 | 69.8 | 69.8 | 88.5 | 69.9 | 78.1 | 67.7 | 38.5 | 49.0 | 77.2 | 63.9 | 69.6 |
| SVM + C | 77.5 | 67.4 | 72.1 | 80.0 | 71.9 | 71.9 | 87.6 | 76.1 | 81.4 | 70.7 | 38.8 | 50.1 | 79.7 | 67.3 | 72.6 |
| SVM + U + C | 86.5 | 75.9 | 80.8 | 66.7 | 48.2 | 55.9 | 74.4 | 68.7 | 71.4 | 78.2 | 73.9 | 73.9 | 77.3 | 69.8 | 73.0 |
| BiLSTM + MFT | 74.9 | 69.4 | 71.9 | 78.8 | 75.4 | 75.4 | 87.3 | 74.6 | 80.5 | 72.8 | 47.5 | 57.5 | 78.6 | 69.7 | 73.7 |
| BiLSTM + RE | 72.3 | 75.3 | 73.8 | 80.1 | 76.9 | 76.9 | 85.7 | 78.7 | 82.1 | 71.1 | 45.2 | 55.2 | 77.3 | 72.9 | 74.8 |
| BiLSTM + IFT | 77.2 | 74.9 | 75.9 | 79.3 | 78.2 | 78.2 | 88.2 | 81.0 | 84.5 | 71.5 | 60.5 | 65.6 | 79.6 | 75.5 | 77.5 |
| Average | 75.7 | 68.8 | 71.9 | 76.5 | 68.4 | 69.5 | 83.3 | 73.9 | 78.3 | 71.8 | 50.2 | 58.3 | 77.3 | 68.1 | 72.1 |

Table 6: Result for Task B: Identification of target of HS. Results are sorted in ascending order based on weighted F1. **U**: word unigram, **C**: char *ngrams*

| Model + Feature | Slander | | | Religion | | | Gender | | | CV | | | Weighted Average | | |
|---|---|---|---|---|---|---|---|---|---|---|---|---|---|---|---|
| | P | R | F1 | P | R | F1 | P | R | F1 | P | R | F1 | P | R | F1 |
| BiLSTM + BFT | 84.9 | 89.3 | 87.0 | 88.9 | 31.2 | 46.2 | 81.8 | 56.9 | 67.1 | 78.6 | 82.7 | 80.6 | 83.1 | 79.9 | 80.5 |
| BiLSTM + MFT | 89.9 | 92.6 | 91.3 | 96.4 | 69.5 | 80.8 | 84.1 | 74.4 | 78.9 | 84.1 | 81.4 | 82.7 | 87.9 | 85.9 | 86.8 |
| SVM + U | 89.2 | 93.1 | 91.1 | 94.2 | 62.9 | 75.5 | 90.2 | 72.8 | 80.6 | 86.5 | 80.5 | 83.4 | 88.9 | 85.5 | 86.8 |
| BiLSTM + RE | 92.8 | 89.6 | 91.2 | 84.9 | 76.6 | 80.6 | 79.87 | 82.77 | 81.3 | 83.3 | 82.0 | 82.6 | 88.2 | 86.1 | 87.1 |
| SVM + U + C | 92.4 | 93.3 | 92.8 | 90.4 | 73.4 | 81.0 | 87.8 | 78.2 | 82.7 | 85.4 | 83.5 | 84.4 | 89.9 | 87.7 | 88.7 |
| BiLSTM + IFT | 93.4 | 92.2 | 92.8 | 92.2 | 76.6 | 83.6 | 85.0 | 79.8 | 82.3 | 88.1 | 82.7 | 85.3 | 90.8 | 87.3 | 88.9 |
| SVM + C | 92.3 | 93.6 | 92.9 | 96.5 | 71.4 | 82.1 | 88.6 | 77.3 | 82.57 | 87.31 | 83.7 | 85.4 | 90.8 | 87.7 | 89.1 |
| Average | 90.7 | 91.9 | 91.3 | 92.9 | 65.9 | 75.7 | 85.3 | 74.6 | 79.4 | 84.7 | 82.4 | 83.5 | 88.5 | 85.7 | 86.8 |

Table 7: Result for Task C: Categorization of HS types. Results are sorted in ascending order based on weighted F1. **U**: word unigram, **C**: char *ngrams*

the best F1 score of 89.1. Although, BiLSTM trained with **IFT** achieved a comparable F1 score of 88.9. And similar with sub-tasks A and B, here also **IFT** proves to be more up to the task compared to two embeddings trained with formal text, **BFT** and **MFT**.
From average recall, we can deduce that, all of the models had the most difficulty correctly identifying *Religion* as the offense type.

# 5. Discussion

## 5.1. Superior performance of Informal Embeddings:

From **Tables 5**, **6**, and **7**, we can observer that informal embeddings (**IFT**) performs significantly better as feature extractors compared to formal word embeddings such as; *BengFastText* (**BFT**) or Multilingual Fast Text (**MFT**). It is noteworthy, as our **IFT** is trained with only 1.47 million informal sentences; which is a fraction of the corpus size large pretrained embeddings like **MFT** and **BFT** were trained with. While the actual cause of this phenomena is left out for future investigations, we present a hypothesis for the cause. **BFT** and **MFT** were trained with much larger *Wikipedia* data, which are *formal* texts and seldom contain grammatical and spelling errors. However, comments on social media platforms are rife with such errors. As our **IFT** was trained with *informal* comments scraped from social media it has *adapted* to texts with mixed dialects as well as grammatical and spelling errors.

## 5.2. Influence of swear words:

Our test set comprised of 15% of out total dataset. We extracted the vocabulary of this test set and identified the swear words. These swear words were divided into two groups; traditional swear (**TS**) and non-traditional swear (**NTS**). We define a swear word as non-traditional if it is not typically used as a swear but can be interpreted as such depending on the context. Figure 1 and 2 shows some examples of both types of swear words. Then we analyzed how our classifier models performed in identifying HS comments that contain a certain type of swear. It is to be noted that a comment might contain both TS and NTS. But, for simplicity of analyzing we omitted these comments in this analysis. **Table 8** depicts the accuracy of our classifiers in detecting each type of swear. From **Table 8**, we can observe that all of the classifiers perform comparatively poorly at detecting HS with NTS.

| **Model+Feature** | $TS_{acc}$ | $NTS_{acc}$ |
|---|---|---|
| SVM + C | 84.27 | 79.53 |
| Bi-LSTM+BFT | 75.71 | 73.31 |
| Bi-LSTM+MFT | 82.17 | 76.16 |
| Bi-LSTM+IFT | 85.08 | 77.97 |

Table 8: Influence of traditional and non traditional swear words in HS detection.

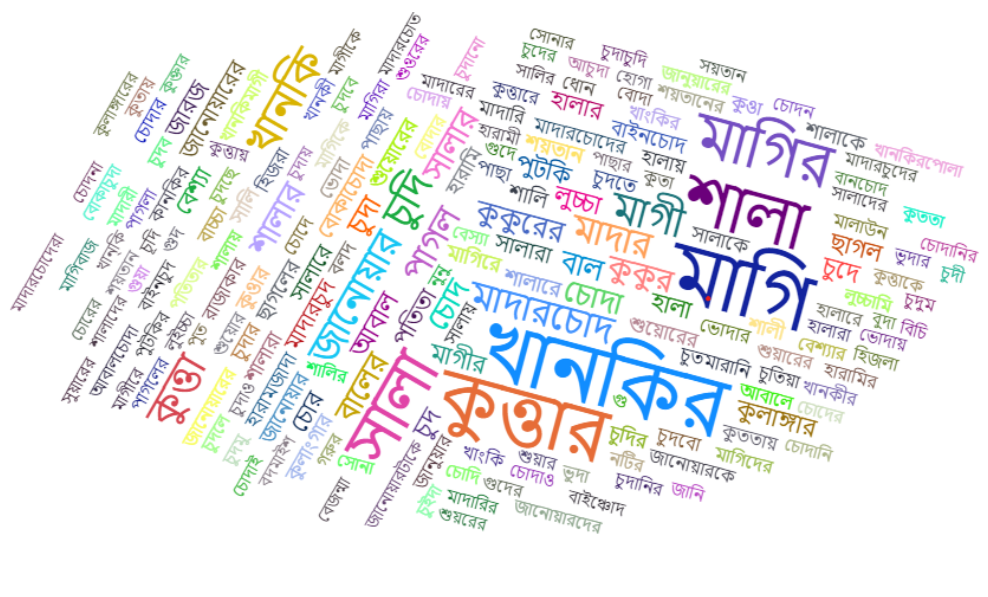

Figure 1: Traditional Swear Word Cloud

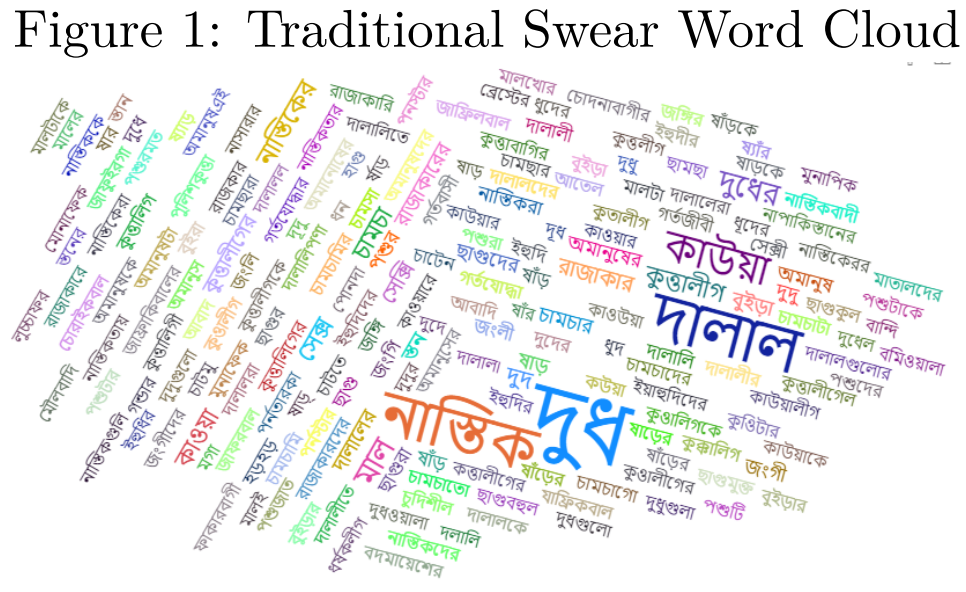

Figure 2: Non Traditional Swear Word Cloud

### 5.3. Limitations of model's performance

**Figures 3** and **4** demonstrates *Venn* diagrams representing the comment overlappings across targets and HS types respectively. It is worthy of noting that, HS types (**Figure 4**) has more overlapping between its classes compared to HS targets (**Figure 3**). However, our classifiers perform poorly at correctly identifying HS targets (**Table 6**) compared to HS types (**Table 7**). One possible explanation for this is that, to identify HS targets classifiers need to be capable of distinguishing between numerous proper and common nouns. Whereas, the classifiers only need to consider certain swear words to correctly identify HS types. Some example of such swear words are depicted in **Figures 1** and **2**.

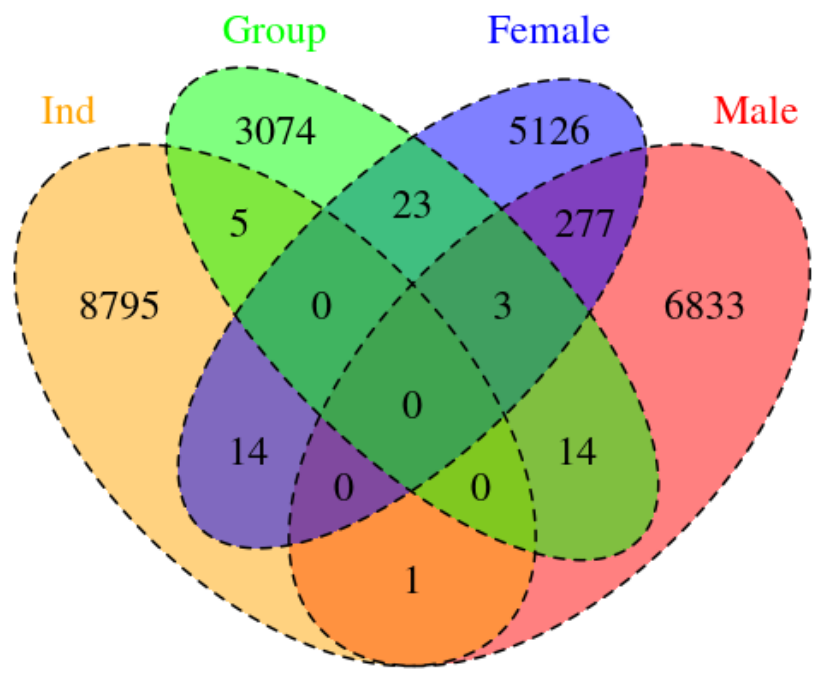

Figure 3: Venn diagram showing comment overlapping across different four HS targets.

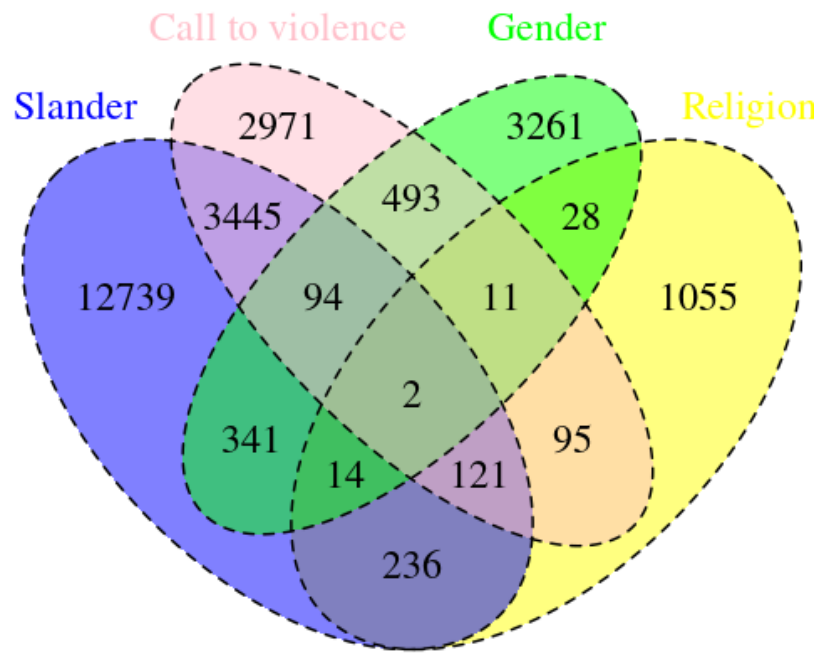

Figure 4: Venn diagram showing HS comments overlapping across different four HS types.

## 6. Conclusion

This paper introduces a manually labelled HS dataset in the Bangla language, obtained from online social media and video streaming sites by crawling comment sections. The resulting dataset contains 50,281 comments systematically labelled to support training machine learning classifiers for three sub-tasks: 1) HS identification, 2) target identification, and 3) HS type identification. We followed various schemes to ensure linguistic diversity and reduce repetitiveness. We conducted several baseline experiments by training machine learning classifiers with traditional count-based features and semantic features. In capturing semantic features from online comments, we found that the language model trained from scratch on *informal* text consistently proves better than larger pretrained language models trained with formal internet texts. In future, data annotation schemes in levels 2 and 3 can be improved as in the current data, a single annotator annotated each comment. In addition, we observed that the annotators had conflicting annotations at level 1 even with our detailed annotation guidelines. In the future, more analyses can be done on conflicting annotations and interpret the mode for better understanding.

# 7. Bibliographical References